\def\year{2020}\relax
\documentclass[letterpaper]{article} 
\usepackage{aaai20}  
\usepackage{times}  
\usepackage{helvet} 
\usepackage{courier}  
\usepackage[hyphens]{url}  
\usepackage{graphicx} 
\usepackage{amsmath,amssymb,bm}
\usepackage{subcaption}
\usepackage{multirow}
\usepackage{graphicx}  
\newcommand{\citet}[1]{\citeauthor{#1} \shortcite{#1}}
\newcommand{\citep}{\cite}
\usepackage[ruled,vlined]{algorithm2e}
\title{Scalable Inference for Nonparametric Hawkes Process Using P\'{o}lya-Gamma Augmentation}
\author{Feng Zhou\textsuperscript{\rm 1,2}, Zhidong Li\textsuperscript{\rm 3}, Xuhui Fan\textsuperscript{\rm 2}, Yang Wang\textsuperscript{\rm 3}, Arcot Sowmya\textsuperscript{\rm 2}, Fang Chen\textsuperscript{\rm 3}\\ 
\textsuperscript{\rm 1}Data61, CSIRO\\
\textsuperscript{\rm 2}University of New South Wales\\
\textsuperscript{\rm 3}University of Technology Sydney\\
}
\begin{document}

\maketitle

\begin{abstract}
In this paper, we consider the sigmoid Gaussian Hawkes process model: the baseline intensity and triggering kernel of Hawkes process are both modeled as the sigmoid transformation of random trajectories drawn from Gaussian processes (GP). By introducing auxiliary latent random variables (branching structure, P\'{o}lya-Gamma random variables and latent marked Poisson processes), the likelihood is converted to two decoupled components with a Gaussian form which allows for an efficient conjugate analytical inference. Using the augmented likelihood, we derive an expectation-maximization (EM) algorithm to obtain the maximum a posteriori (MAP) estimate. Furthermore, we extend the EM algorithm to an efficient approximate Bayesian inference algorithm: mean-field variational inference. We demonstrate the performance of two algorithms on simulated fictitious data. Experiments on real data show that our proposed inference algorithms can recover well the underlying prompting characteristics efficiently. 
\end{abstract}

\section{Introduction}
\subsubsection{Background}
The Hawkes process is one important class of point processes which can be utilized to model the \textit{self-exciting} phenomenon in numerous application domains, e.g. crime \citep{liu2018exploiting}, ecosystem \citep{gupta2018discrete}, transportation \citep{du2016recurrent} and social network \citep{pinto2015trend}. An important statistic of point processes is the conditional intensity: the probability of one event occurring in an infinitesimal time interval given history. Specifically, the conditional intensity of Hawkes process is 
\begin{equation}
\lambda(t)=\mu(t)+\int_0^t\phi(t-s)d\mathbb{N}(s)=\mu(t)+\sum_{t_i<t}\phi(t-t_i),
\label{eq1}
\end{equation}
where $\mu(t)>0$ is the baseline intensity, $\{t_i\}$ are timestamps of events occurring before $t$, $\mathbb{N}(t)$ is the corresponding counting process and $\phi(\tau)>0$ where $\tau=t-t_i$ is the triggering kernel. The summation of triggering kernels explains the nature of self-excitation: events occurring in the past intensify the rate of occurrence in the future. 

The classic Hawkes process is supposed to be in a parametric form: the baseline intensity $\mu(t)$ is assumed to be a constant with triggering kernel $\phi(\tau)$ being a parametric function, e.g. exponential decay or power law decay function. However, in reality, the actual exogenous rate $\mu(t)$ can change over time due to the varying exterior context; the actual endogenous rate capturing how
previous events trigger posterior ones, which is modeled by $\phi(\tau)$, can be rather complex among different applications. For example, the exogenous rate of civilian deaths due to insurgent activity is changing over time \citep{lewis2011nonparametric} and the prompting effect of vehicle collision decays periodically and oscillatingly \citep{zhou2018refined}. Obviously, the models based on classic Hawkes process tend to be oversimplified or even incapable of capturing the ground truth in numerous scenarios. Therefore, it is remarkably necessary to estimate the exogenous and endogenous dynamics in a data-driven nonparametric approach. 

A wide variety of nonparametric estimation approaches of Hawkes process have been largely investigated over past few years. From frequentist nonparametric perspective, \citet{marsan2008extending} proposed to estimate the triggering kernel modeled as a histogram function with an EM algorithm and \citet{lewis2011nonparametric} extended this method by introducing a smooth regularizer and performed estimation by solving a Euler-Lagrange equation, \citet{zhou2013learning} further extended this algorithm to multivariate Hawkes process; \citet{bacry2016first} provided an estimation approach that is based on the solution of a Wiener-Hopf equation relating the triggering kernel with the second order statistics; \citet{eichler2017graphical,reynaud2010adaptive} attempted to minimize a quadratic contrast function with a grid based triggering kernel. From Bayesian nonparametric perspective, most related works are based on Gaussian-Cox processes: the Poisson process with a stochastic intensity modulated by GP. To guarantee the non–negativity of intensity, trajectories drawn from a GP prior need to be squashed by a link function. For example, a log-Gaussian intensity is utilized by \citet{moller1998log,samo2015scalable}; \citet{adams2009tractable} proposed a sigmoid-GP intensity and a tractable Markov chain Monte Carlo (MCMC) algorithm. \citet{lloyd2015variational} developed a variational Gaussian approximation algorithm with a square link function. As far as we know, only a small amount of works attempted to infer Hawkes process with GP prior since Hawkes process is more complicated than Poisson. For example, \citet{zhou2018refined} added an extra GP regression step into the EM algorithm by \citet{marsan2008extending} to achieve smoothness and facilitate choice of hyperparameters; \citet{zhou2019efficient} extended the variational inference algorithm by \citet{lloyd2015variational} to Hawkes process. 

\subsubsection{Issues}

All GP modulated intensity models mentioned above have the same issue: \textbf{1}) due to the existence of link function, the likelihood of GP variables is non-conjugate to the prior resulting in a non-Gaussian posterior. The non-conjugacy leads to a complicated and time-consuming inference procedure. \textbf{2}) Furthermore, in Hawkes process, the exogenous component (baseline intensity) and the endogenous component (triggering kernel) are coupled in the likelihood, which further hampers the tractability of inference. 

\subsubsection{Targets}

To circumvent these problems, we augment the likelihood with auxiliary latent random variables: \textit{branching structure}, \textit{P\'{o}lya-Gamma random variables} and \textit{latent marked Poisson processes}. The branching structure of Hawkes process is introduced to decouple $\mu(t)$ and $\phi(\tau)$ to two independent components in the likelihood; inspired by \citet{polson2013bayesian} and \citet{donner2018efficient}, we use a sigmoid link function in the model and convert the sigmoid to an infinite mixture of Gaussians involving P\'{o}lya-Gamma random variables; the latent marked Poisson processes are augmented to linearize the exponential integral term in likelihood. By augmenting the likelihood in such a way, the likelihood becomes conjugate to the GP prior. With these latent random variables, we use the augmented likelihood to construct two efficient analytical iterative algorithms. The first one is an EM algorithm to obtain the MAP estimate; furthermore, we extend the EM algorithm to a mean-field variational inference algorithm that is slightly slower but can provide uncertainty with a distribution estimation rather than point estimation. It is worth noting that the na\"ive implementations of both algorithms are time-consuming. To improve efficiency remarkably, the sparse GP approximation \citep{titsias2009variational} is introduced. 

\subsubsection{Contributions} 

Specifically, we make the following contributions:

\textbf{1.} An EM algorithm and a mean-field varitional inference algorithm are derived for Bayesian nonparametric Hawkes process: the baseline intensity and triggering kernel are both sigmoid-GP rates. 

\textbf{2.} The original Hawkes process likelihood is converted to two decoupled factors which are conjugate to GP priors by augmenting the branching structure, P\'{o}lya-Gamma random variables and latent marked Poisson processes. This results in simple and efficient EM and varitional inference algorithms with explicit expressions. 

\textbf{3.} The complexity is further reduced to be almost linear due to the utilization of sparse GP approximation. 

\section{Sigmoid GP Hawkes Process}
A Hawkes process is a stochastic process whose realization is a sequence of timestamps $D=\{t_i \}_{i=1}^N\in[0,T]$. Here, $t_i$ stands for the occurrence time of $i$-th event with $T$ being the observation window. The conditional intensity of Hawkes process is already provided in Eq.\eqref{eq1}. Given $\mu(t)$ and $\phi(\tau)$, the Hawkes process likelihood \citep{daley2003introduction} is
\begin{equation}
\label{eq2}
\begin{aligned}
p(D|\mu(t),&\phi(\tau))=\prod_{i=1}^{N}\left[\mu(t_i)+\sum_{t_j<t_i}\phi(t_i-t_j)\right]\cdot\\
&\exp\left(-\int_{T}\left(\mu(t)+\sum_{t_i<t}\phi(t-t_i)\right)dt\right).
\end{aligned}
\end{equation}

We propose a GP based Bayesian nonparametric Hawkes process model: sigmoid-GP Hawkes process (SGPHP) whose baseline intensity and triggering kernel are functions drawn from a GP prior, passed through a sigmoid link function and scaled by an upper-bound to guarantee the non-negativity: $\mu(t)=\lambda^*_{\mu}\sigma(f(t)),\phi(\tau)=\lambda^*_{\phi}\sigma(g(\tau))$ where $\sigma(\cdot)$ is the sigmoid function, $f$ and $g$ are two functions drawn from the corresponding GP priors, $\lambda^*_{\mu}$ and $\lambda^*_{\phi}$ are the upper-bounds of $\mu(t)$ and $\phi(\tau)$. 

In a na\"ive Bayesian framework, the inference of posterior of $\mu(t)$ and $\phi(\tau)$ is non-trivial because of 1) the doubly-intractable problem introduced by \citet{adams2009tractable} caused by intractable integrals in the numerator and denominator; 2) the posterior being non-Gaussian. However, as we can see later, these two problems can be circumvented by augmenting the likelihood with auxiliary latent random variables. The sigmoid link function is chosen since it can be transformed to infinite mixture of Gaussians; consequently, the augmented likelihood is in a conjugate form allowing for more efficient EM and varitional inference with explicit expressions. 

\section{Likelihood Augmentation}

\subsection{Augmenting Branching Structure}

In the above likelihood, the coupling of $\mu(t)$ and $\phi(\tau)$ in the products term leads to inference difficulty. A well-known decoupling method is to incorporate the \textit{branching structure} of Hawkes process \citep{marsan2008extending,zhou2013learning}. The branching structure $\mathbf{X}$ is a triangular matrix with Bernoulli variables $x_{ij}$ indicating whether the $i$-th event is triggered by itself or a previous event $j$. 
\begin{equation*}
    \begin{split}
    &x_{ii}=\left\{
                        \begin{aligned}
                        &1 \ \ \text{if event $i$ is a background event}\\
                        &0 \ \ \ \ \ \ \ \ \ \ \ \ \ \ \text{otherwise}\\
                        \end{aligned}
                    \right.\\
    &x_{ij}=\left\{
                        \begin{aligned}
                        &1 \ \ \text{if event $i$ is caused by event $j$}\\
                        &0 \ \ \ \ \ \ \ \ \ \ \ \ \ \ \text{otherwise}\\
                        \end{aligned}
                    \right.
    \end{split}
\end{equation*}
The joint likelihood with branching structure is 
\begin{equation}
\label{eq3}
\begin{aligned}
&p(D,\mathbf{X}|\mu(t),\phi(\tau))=\underbrace{\prod_{i=1}^N\mu(t_i)^{x_{ii}}\exp{\left(-\int_T\mu(t)dt\right)}}\cdot\\
&\underbrace{\prod_{i=2}^N\prod_{j=1}^{i-1}\phi(t_i-t_j)^{x_{ij}}\prod_{i=1}^N\exp{\left(-\int_{T_\phi}\phi(\tau)d\tau\right)}},
\end{aligned}
\end{equation}
where $T_\phi$ is the support of triggering kernel, $\mu(t)=\lambda^*_{\mu}\sigma(f(t))$, $\phi(\tau)=\lambda^*_{\phi}\sigma(g(\tau))$. After introducing the branching structure, the joint likelihood is decoupled to two independent factors. 

\subsection{Transformation of Sigmoid Function}

We utilize a remarkable representation discovered in the literature of Bayesian inference for logistic regression \citep{polson2013bayesian} in recent years. Surprisingly, the sigmoid function is redefined as a Gaussian representation
\begin{equation}
\label{eq4}
\sigma(z)=\int_0^\infty e^{h(\omega,z)}p_{\text{PG}}(\omega|1,0)d\omega,
\end{equation}
where $h(\omega,z)=z/2-z^2\omega/2-\log2$, $p_{\text{PG}}(\omega|1,0)$ is the P\'{o}lya-Gamma distribution with $\omega\in \mathbb{R}^+$. The derivation is shown in the Appendix~\ref{app1}. 

Using Eq.\eqref{eq4}, the products of observations $\sigma(f(t_i))$ and $\sigma(g(\tau_{ij}))$ ($\tau_{ij}=t_i-t_j$) in
the likelihood Eq.\eqref{eq3} are transformed into a Gaussian form. It is worth noting that we do not need to know the exact form of P\'{o}lya-Gamma distribution but only its first order moment. 

\subsection{Transformation of Exponential Integral} 

Here, we only discuss the baseline intensity part. All derivation in the triggering kernel part is same as the baseline intensity part except some notations. Utilizing Eq.\eqref{eq4} and the sigmoid property $\sigma(z)=1-\sigma(-z)$, the exponential integral in the likelihood Eq.\eqref{eq3} can be rewritten as
\begin{equation}
\label{eq5}
\begin{aligned}
&\exp{\left(-\int_T\lambda^*_{\mu}\sigma(f(t))dt\right)}=\\
&\exp{\left(-\int_T\int_{\mathbb{R}^+}\left(1-e^{h(\omega_{\mu},-f(t))}\right)\lambda^*_{\mu}p_{\text{PG}}(\omega_{\mu}|1,0)d\omega_{\mu} dt\right)}
\end{aligned}
\end{equation}
According to Campbell's theorem \citep{kingman2005p}, the right hand side of Eq.\eqref{eq5} is a characteristic functional of a marked Poisson process, so we can rewrite it as
\begin{equation}
\label{eq6}
\begin{aligned}
\exp{\left(-\int_T\lambda^*_{\mu}\sigma(f(t))dt\right)}=\mathbb{E}_{p_{\lambda_\mu}}\left[\prod_{(\omega_{\mu},t)\in\Pi_{\mu}}e^{h(\omega_{\mu},-f(t))}\right],
\end{aligned}
\end{equation}
where $\Pi_{\mu}=\{({\omega_{\mu}}_m,t_m)\}_{m=1}^{M_\mu}$ denotes a random realization of a marked Poisson process and ${p_{\lambda_\mu}}$ is the probability measure of the marked Poisson process $\Pi_{\mu}$ with intensity $\lambda_{\mu}(t,\omega_{\mu})=\lambda^*_{\mu}p_{\text{PG}}(\omega_{\mu}|1,0)$. The events $\{t_m\}_{m=1}^{M_\mu}$ follow a Poisson process with rate $\lambda^*_{\mu}$ and the latent P\'{o}lya-Gamma variable ${\omega_{\mu}}_m$ denotes the independent \textit{mark} at each location $t_m$. The detailed derivation can be found in the Appendix ~\ref{app2}. 

\subsection{Augmented Likelihood} 

Substituting Eq.\eqref{eq4} and Eq.\eqref{eq6} into Eq.\eqref{eq3}, we obtain 

\textbf{1.} the augmented joint likelihood of baseline intensity part (derivation in Appendix~\ref{app3})
\begin{equation}
\label{eq7}
\begin{aligned}
p(D,\Pi_{\mu},\bm{\omega}_{ii},\mathbf{X}_{ii}|&\lambda^*_{\mu},f)=\prod_{i=1}^N\left(\lambda_{\mu}(t_i,\omega_{ii})e^{h(\omega_{ii},f(t_i))}\right)^{x_{ii}}\\
&\cdot p_{\lambda_\mu}(\Pi_\mu|\lambda^*_{\mu})\prod_{(\omega_{\mu},t)\in \Pi_\mu}e^{h(\omega_\mu,-f(t))}
\end{aligned}
\end{equation}
with $\bm{\omega}_{ii}$ denoting a vector of $\omega_{ii}$ on each $t_i$ and $\mathbf{X}_{ii}$ being the diagonal of branching structure $\mathbf{X}$; 

\textbf{2.} and the augmented joint likelihood of triggering kernel part (derivation in Appendix~\ref{app3}) 
\begin{equation}
\label{eq8}
\begin{aligned}
&p(D,\{\Pi_{\phi_i}\}_{i=1}^N,\bm{\omega}_{ij},\mathbf{X}_{ij}|\lambda^*_{\phi},g)=\\
&\prod_{i=2}^N\prod_{j=1}^{i-1}\left(\lambda_{\phi}(\tau_{ij},\omega_{ij})e^{h(\omega_{ij},g(\tau_{ij}))}\right)^{x_{ij}}\cdot\\
&\prod_{i=1}^N\left[p_{\lambda_\phi}(\Pi_{\phi_i}|\lambda^*_{\phi})\prod_{(\omega_{\phi},\tau)\in \Pi_{\phi_i}}e^{h(\omega_\phi,-g(\tau))}\right],
\end{aligned}
\end{equation}
where ${p_{\lambda_\phi}}$ is the probability measure of the corresponding latent marked Poisson process $\Pi_{\phi_i}=\{({\omega_\phi}_m, \tau_m)\}_{m=1}^{M_{\phi_i}}$ with intensity $\lambda_{\phi}(\tau,\omega_{\phi})=\lambda^*_{\phi}p_{\text{PG}}(\omega_{\phi}|1,0)$, $\bm{\omega}_{ij}$ denotes the vector of $\omega_{ij}$ on each $\tau_{ij}$ and $\mathbf{X}_{ij}$ is the entries off the diagonal of branching structure. It is worth noting that there exists $N$ independent latent marked Poisson processes because of the exponential integral product term in Eq.\eqref{eq3}. 

The motivation of augmenting  auxiliary  latent  random  variables should now be clear. The augmented representation of likelihood contains the GP variables $f(\cdot)$ and $g(\cdot)$ only linearly and quadratically in the exponents and is thus conjugate to the GP prior.

\section{EM Algorithm}

With the original likelihood Eq.\eqref{eq2} and GP priors $\mathcal{GP}(f)$ and $\mathcal{GP}(g)$ (symmetric prior $\mathcal{GP}(\cdot|0,K_{\cdot})$), the log-posterior corresponds to a penalized log-likelihood. As discussed by \citet{donner2019bayesian} and \citet{rasmussen2003gaussian} for GP models with likelihood depending on finite inputs only, the regularizer is given by the squared reproducing kernel Hilbert space (RKHS) norm corresponding to the GP kernel. Therefore, we obtain
\begin{equation}
\label{eq9}
\begin{aligned}
\hat{\lambda}^*_{\mu},\hat{f},\hat{\lambda}^*_{\phi},\hat{g}=&\text{argmax}\bigg\{\log p(D|\lambda^*_{\mu},f,\lambda^*_{\phi},g)\\
&-\frac{1}{2}\Vert f\Vert^2_{\mathcal{H}_{k_f}}-\frac{1}{2}\Vert g\Vert^2_{\mathcal{H}_{k_g}}\bigg\},
\end{aligned}
\end{equation}
where $\hat{\lambda}^*_{\mu},\hat{f},\hat{\lambda}^*_{\phi},\hat{g}$ are the MAP estimates, $\Vert \cdot\Vert^2_{\mathcal{H}_{k}}$ is the squared RKHS norm with kernel $k$. The regularizer is the functional counterpart of log Gaussian prior. Instead of performing direct optimization, we propose an EM algorithm with the augmented auxiliary variables. Specifically, we propose a lower-bound of the log-posterior
\begin{equation}
\label{eq10}
\begin{aligned}
&\mathcal{Q}((\lambda^*_{\mu},f,\lambda^*_{\phi},g)|(\lambda^*_{\mu},f,\lambda^*_{\phi},g)_{\text{old}})=\\
&\mathbb{E}\left[\log p(D,\Pi_{\mu},\bm{\omega}_{ii},\{\Pi_{\phi_i}\}_{i=1}^N,\bm{\omega}_{ij},\mathbf{X}|\lambda^*_{\mu},f,\lambda^*_{\phi},g)\right]\\
&-\frac{1}{2}\Vert f\Vert^2_{\mathcal{H}_{k_f}}-\frac{1}{2}\Vert g\Vert^2_{\mathcal{H}_{k_g}}, 
\end{aligned}
\end{equation}
with $\mathbb{E}$ over $P(\Pi_{\mu},\bm{\omega}_{ii},\{\Pi_{\phi_i}\}_{i=1}^N,\bm{\omega}_{ij},\mathbf{X}|(\lambda^*_{\mu},f,\lambda^*_{\phi},g)_{\text{old}})$. 

Because of auxiliary variables augmentation, the GP variables are in a quadratic form in the lower-bound, which results in an analytical solution in the M step. 

\subsection{E Step}
In the E step, we first derive the \textit{conditional density} $P(\Pi_{\mu},\bm{\omega}_{ii},\{\Pi_{\phi_i}\}_{i=1}^N,\bm{\omega}_{ij},\mathbf{X}|(\lambda^*_{\mu},f,\lambda^*_{\phi},g)_{\text{old}})$ and then compute the \textit{lower-bound} $\mathcal{Q}$. 

\subsubsection{Conditional density}

The conditional density of $\Pi_{\mu}$, $\bm{\omega}_{ii}$, $\{\Pi_{\phi_i}\}_{i=1}^N$, $\bm{\omega}_{ij}$, $\mathbf{X}$ given $(\lambda^*_{\mu},f,\lambda^*_{\phi},g)_{\text{old}}$ can be factorized and obtained from Eq.\eqref{eq7} and \eqref{eq8}. More specifically, we provide details of these factors. 


\textbf{1.} The conditional distributions of P\'{o}lya-Gamma variables $\bm{\omega}_{ii}$ and $\bm{\omega}_{ij}$ depend on the function values $f_{\text{old}}$ and $g_{\text{old}}$ at $t_i$ and $\tau_{ij}$
\begin{equation}
\label{eq11}
\begin{aligned}
p(\bm{\omega}_{ii}|\mathbf{f}_{\text{old}})&=\prod_{i=1}^Np_{\text{PG}}(\omega_{ii}|1,f_{\text{old}}(t_i))\\
p(\bm{\omega}_{ij}|\mathbf{g}_{\text{old}})&=\prod_{i=2}^N\prod_{j=1}^{i-1}p_{\text{PG}}(\omega_{ij}|1,g_{\text{old}}(\tau_{ij})),
\end{aligned}
\end{equation}
where we marginalize out $\mathbf{X}$ and utilize the tilted P\'{o}lya-Gamma distribution $p_{\text{PG}}(\omega|b,c)\propto e^{-c^2\omega/2}p_{\text{PG}}(\omega|b,0)$ with the first order moment being $\mathbb{E}[\omega]=\frac{b}{2c}\tanh{\frac{c}{2}}$ \citep{polson2013bayesian}. 


\textbf{2.} The conditional density of $\Pi_\mu$ depends on $f_{\text{old}}$ and ${\lambda_\mu^*}_{\text{old}}$
\begin{equation}
\label{eq12}
\begin{aligned}
&p(\Pi_\mu|f_{\text{old}},{\lambda_\mu^*}_{\text{old}})=\\
&\frac{p_{\lambda_\mu}(\Pi_\mu|{\lambda_\mu^*}_{\text{old}})\prod_{(\omega_\mu,t)\in\Pi_\mu}e^{h(\omega_\mu,-f_{\text{old}}(t))}}{\int p_{\lambda_\mu}(\Pi_\mu|{\lambda_\mu^*}_{\text{old}})\prod_{(\omega_\mu,t)\in\Pi_\mu}{e^{h(\omega_\mu,-f_{\text{old}}(t))}}d\Pi_\mu}.
\end{aligned}
\end{equation}

Using Eq.\eqref{eq5} and \eqref{eq6} to convert the denominator, Eq.\eqref{eq12} can be rewritten as
\begin{equation*}
\begin{aligned}
&p(\Pi_\mu|f_{\text{old}},{\lambda_\mu^*}_{\text{old}})=\\
&\frac{p_{\lambda_\mu}(\Pi_\mu|{\lambda_\mu^*}_{\text{old}})\prod_{(\omega_\mu,t)\in\Pi_\mu}e^{h(\omega_\mu,-f_{\text{old}}(t))}}{\exp{(-\iint(1-e^{h(\omega_\mu,-f_{\text{old}}(t))}){\lambda_\mu^*}_{\text{old}}p_{\text{PG}}(\omega_\mu|1,0)d\omega_\mu dt)}}=\\
&\prod_{(\omega_\mu,t)\in\Pi_\mu}\left(e^{h(\omega_\mu,-f_{\text{old}}(t))}{\lambda_\mu^*}_{\text{old}}p_{\text{PG}}(\omega_\mu|1,0)\right)\cdot\\
&\exp{\left(-\iint e^{h(\omega_\mu,-f_{\text{old}}(t))}{\lambda_\mu^*}_{\text{old}}p_{\text{PG}}(\omega_\mu|1,0)d\omega_\mu dt\right)}
\end{aligned}
\end{equation*}

It is straightforward to see the above conditional distribution is in the likelihood form of a marked Poisson process with intensity function
\begin{equation}
\label{eq13}
\begin{aligned}
\Lambda_\mu(t,\omega_\mu)&=e^{h(\omega_\mu,-f_{\text{old}}(t))}{\lambda_\mu^*}_{\text{old}}p_{\text{PG}}(\omega_\mu|1,0)\\
&={\lambda_\mu^*}_{\text{old}}\sigma(-f_{\text{old}}(t))p_{\text{PG}}(\omega_\mu|1,f_{\text{old}}(t)). 
\end{aligned}
\end{equation}

The derivation of conditional distribution of $\Pi_{\phi_i}$ is same as $\Pi_\mu$ with the corresponding subscripts being replaced. It is worth noting that there exists $N$ independent marked Poisson processes $\{\Pi_{\phi_i}\}_{i=1}^N$ with the same intensity function
\begin{equation}
\label{eq14}
\Lambda_\phi(\tau,\omega_\phi)={\lambda_\phi^*}_{\text{old}}\sigma(-g_{\text{old}}(\tau))p_{\text{PG}}(\omega_\phi|1,g_{\text{old}}(\tau)). 
\end{equation}


\textbf{3.} Combining Eq.\eqref{eq7} and \eqref{eq8} and marginalizing out $\bm{\omega}_{ii}$ and $\bm{\omega}_{ij}$, we obtain the conditional distribution of $\mathbf{X}$ 
\begin{equation*}
\begin{aligned}
&p(\mathbf{X}|(\lambda^*_{\mu},f,\lambda^*_{\phi},g)_{\text{old}})\propto\prod_{i=1}^N\left(\mu_{\text{old}}(t_i)\right)^{x_{ii}}\prod_{i=2}^N\prod_{j=1}^{i-1}\left(\phi_{\text{old}}(\tau_{ij})\right)^{x_{ij}},
\end{aligned}
\end{equation*}
with $\mu_{\text{old}}(t_i)={\lambda_\mu^*}_{\text{old}}\sigma(f_{\text{old}}(t_i))$ and $\phi_{\text{old}}(\tau_{ij})={\lambda_\phi^*}_{\text{old}}\sigma(g_{\text{old}}(\tau_{ij}))$. This is a multinomial distribution with
\begin{equation}
\label{eq15}
\begin{aligned}
p(x_{ii}=1)&=\frac{\mu_{\text{old}}(t_i)}{\mu_{\text{old}}(t_i)+\sum_{j=1}^{i-1}\phi_{\text{old}}(\tau_{ij})}\\
p(x_{ij}=1)&=\frac{\phi_{\text{old}}(\tau_{ij})}{\mu_{\text{old}}(t_i)+\sum_{j=1}^{i-1}\phi_{\text{old}}(\tau_{ij})}
\end{aligned}
\end{equation}
which is a well-known result in \citet{lewis2011nonparametric} and \citet{zhou2013learning}. 

\subsubsection{Lower-bound of log-posterior} 

Given those conditional densities above, we can compute the lower-bound $\mathcal{Q}$. The expectation of log-likelihood (ELL) term in Eq.\eqref{eq10} can be rewritten as the summation of baseline intensity part and triggering kernel part. The ELL of baseline intensity part is 
\begin{equation}
\label{eq16}
\begin{aligned}
&\text{ELL}_{\mu}(\lambda_\mu^*,f)\\
&=\mathbb{E}_{P(\Pi_{\mu},\bm{\omega}_{ii},\mathbf{X}_{ii}|(\lambda^*_{\mu},f,\lambda^*_{\phi},g)_{\text{old}})}\left[\log p(D,\Pi_{\mu},\bm{\omega}_{ii},\mathbf{X}_{ii}|\lambda^*_{\mu},f)\right]\\
&=-\frac{1}{2}\int_T A_{\mu}(t)f^2(t)dt+\int_T B_{\mu}(t)f(t)dt-\lambda_{\mu}^*T\\
&\left(\sum_{i=1}^N\mathbb{E}(x_{ii})+\iint\Lambda_\mu(t,\omega_\mu)d\omega_\mu dt\right)\log\lambda_{\mu}^*
\end{aligned}
\end{equation}
where
\begin{equation*}
\begin{aligned}
&A_{\mu}(t)=\sum_{i=1}^N\mathbb{E}[\omega_{ii}]\mathbb{E}[x_{ii}]\delta(t-t_i)+\int_0^\infty \omega_{\mu}\Lambda_{\mu}(t,\omega_{\mu})d\omega_{\mu}\\
&B_{\mu}(t)=\frac{1}{2}\sum_{i=1}^N\mathbb{E}[x_{ii}]\delta(t-t_i)-\frac{1}{2}\int_0^\infty\Lambda_{\mu}(t,\omega_{\mu})d\omega_{\mu},
\end{aligned}
\end{equation*}
with $\mathbb{E}$ over $P(\omega_{ii}|f_{\text{old}}(t_i))$ or $P(x_{ii}|(\lambda^*_{\mu},f,\lambda^*_{\phi},g)_{\text{old}})$. 

Similarly, the ELL of triggering kernel part is written as 
\begin{equation}
\label{eq17}
\begin{aligned}
&\text{ELL}_{\phi}(\lambda_\phi^*,g)=\\
&\mathbb{E}_{P(\{\Pi_{\phi_i}\},\bm{\omega}_{ij},\mathbf{X}_{ij}|(\lambda^*_{\mu},f,\lambda^*_{\phi},g)_{\text{old}})}\left[\log p(D,\{\Pi_{\phi_i}\},\bm{\omega}_{ij},\mathbf{X}_{ij}|\lambda^*_{\phi},g)\right]\\
&=-\frac{1}{2}\int_{T_\phi} A_{\phi}(\tau)g^2(\tau)d\tau+\int_{T_\phi} B_{\phi}(\tau)g(\tau)d\tau-N\lambda_{\phi}^*T_\phi\\
&\left(\sum_{i=2}^N\sum_{j=1}^{i-1}\mathbb{E}(x_{ij})+N\iint\Lambda_\phi(\tau,\omega_\phi)d\omega_\phi d\tau\right)\log\lambda_{\phi}^*
\end{aligned}
\end{equation}
where
\begin{equation*}
\begin{aligned}
A_{\phi}(\tau)=&\sum_{i,j}\mathbb{E}[\omega_{ij}]\mathbb{E}[x_{ij}]\delta(\tau-\tau_{ij})+N\int_0^\infty \omega_{\phi}\Lambda_{\phi}(\tau,\omega_{\phi})d\omega_{\phi}\\
B_{\phi}(\tau)=&\frac{1}{2}\sum_{i,j}\mathbb{E}[x_{ij}]\delta(\tau-\tau_{ij})-\frac{N}{2}\int_0^\infty\Lambda_{\phi}(\tau,\omega_{\phi})d\omega_{\phi},
\end{aligned}
\end{equation*}
with $\mathbb{E}$ over $P(\omega_{ij}|g_{\text{old}}(\tau_{ij}))$ or $P(x_{ij}|(\lambda^*_{\mu},f,\lambda^*_{\phi},g)_{\text{old}})$.

However, the ELL is intractable for general GP priors by the fact that the ELL is a functional. To circumvent the problem, we utilize the sparse GP approximation to introduce some inducing points. $f$ and $g$ are supposed to be dependent on their corresponding inducing points $\{t_s\}_{s=1}^{S_\mu}$ and $\{\tau_s\}_{s=1}^{S_\phi}$; the function values of $f$ and $g$ at these inducing points are $\mathbf{f}_{t_s}$ and $\mathbf{g}_{\tau_s}$. Given a sample $\mathbf{f}_{t_s}$ and $\mathbf{g}_{\tau_s}$, $f(t)$ and $g(\tau)$ in Eq.\eqref{eq16} and \eqref{eq17} are assumed to be the posterior mean functions
\begin{equation}
\label{eq18}
f(t)=\mathbf{k}_{t_s t}^T\mathbf{K}_{t_s t_s}^{-1}\mathbf{f}_{t_s},\ \  g(\tau)=\mathbf{k}_{\tau_s \tau}^T\mathbf{K}_{\tau_s \tau_s}^{-1}\mathbf{g}_{\tau_s},
\end{equation}
with $\mathbf{k}_{t t_s}^T$ and $\mathbf{k}_{\tau \tau_s}^T$ being the kernel vector w.r.t. the observations and inducing points while $\mathbf{K}_{t_s t_s}$ and $\mathbf{K}_{\tau_s \tau_s}$ being w.r.t. inducing points only. 

Substituting Eq.\eqref{eq18} to Eq.\eqref{eq16} and \eqref{eq17}, we can obtain 
\begin{equation}
\label{eq19}
\begin{aligned}
&\mathcal{Q}((\lambda_{\mu}^*,\mathbf{f}_{t_s},\lambda_{\phi}^*,\mathbf{g}_{\tau_s})|(\lambda_{\mu}^*,\mathbf{f}_{t_s},\lambda_{\phi}^*,\mathbf{g}_{\tau_s})_{\text{old}})=\text{ELL}_{\mu}(\lambda_{\mu}^*,\mathbf{f}_{t_s})\\
&+\text{ELL}_{\phi}(\lambda_{\phi}^*,\mathbf{g}_{\tau_s})-\frac{1}{2}\mathbf{f}_{t_s}^T\mathbf{K}_{t_s t_s}^{-1}\mathbf{f}_{t_s}-\frac{1}{2}\mathbf{g}_{\tau_s}^T\mathbf{K}_{\tau_s \tau_s}^{-1}\mathbf{g}_{\tau_s}
\end{aligned}
\end{equation}

\subsection{M Step}

In the M step, we maximize the lower-bound $\mathcal{Q}$. The optimal parameters $\hat{\lambda}^*_{\mu},\hat{\mathbf{f}}_{t_s},\hat{\lambda}^*_{\phi},\hat{\mathbf{g}}_{\tau_s}$ can be obtained by setting the gradient of Eq.\eqref{eq19} to zero. Due to auxiliary variables augmentation, we have analytical solutions
\begin{equation}
\label{eq20}
\begin{aligned}
\hat{\lambda}^*_{\mu}&=\left(\sum_{i=1}^N p_{ii}+M_{\mu}\right)/T\\
\hat{\lambda}^*_{\phi}&=\left(\sum_{i=2}^N\sum_{j=1}^{i-1} p_{ij}+NM_{\phi}\right)/(NT_{\phi})\\
\hat{\mathbf{f}}_{t_s}&=\mathbf{\Sigma}_{t_s}\mathbf{K}_{t_s t_s}^{-1}\int_T B_\mu(t)\mathbf{k}_{t_s t}dt\\
\hat{\mathbf{g}}_{\tau_s}&=\mathbf{\Sigma}_{\tau_s}\mathbf{K}_{\tau_s \tau_s}^{-1}\int_{T_\phi} B_\phi(\tau)\mathbf{k}_{\tau_s \tau}d\tau
\end{aligned}
\end{equation}
where $\mathbf{\Sigma}_{t_s}=\left[\mathbf{K}_{t_s t_s}^{-1}\int A_\mu(t)\mathbf{k}_{t_s t}\mathbf{k}_{t_s t}^Tdt\mathbf{K}_{t_s t_s}^{-1}+\mathbf{K}_{t_s t_s}^{-1}\right]^{-1}$, $\mathbf{\Sigma}_{\tau_s}=\left[\mathbf{K}_{\tau_s \tau_s}^{-1}\int A_\phi(\tau)\mathbf{k}_{\tau_s \tau}\mathbf{k}_{\tau_s \tau}^Td\tau\mathbf{K}_{\tau_s \tau_s}^{-1}+\mathbf{K}_{\tau_s \tau_s}^{-1}\right]^{-1}$, $M_{\mu}=\iint\Lambda_\mu(t,\omega_\mu)d\omega_\mu dt$, $M_{\phi}=\iint\Lambda_\phi(\tau,\omega_\phi)d\omega_\phi d\tau$. All intractable integrals can be solved by Gaussian quadrature. 

\subsection{Complexity}

Another advantage of sparse GP approximation is the complexity of matrix inversion is fixed to $\mathcal{O}(S_\mu^3+S_\phi^3)$ where $S_\mu$ (\text{or} $S_\phi$) $\ll N$. This results in a complexity scaling almost \textbf{linearly} with data size: $\mathcal{O}(NL)$ where $L=\int_{T_\phi}\frac{\mu(t)}{1-\int\phi(\tau)d\tau}dt\ll N$ due to the sparsity of expectation of branching structure: previous points that are more than $T_\phi$ far away from event $i$ have no influence on event $i$ ($\mathbb{E}[x_{ij}]=0$). 

\subsection{Hyperparameters}
Throughout this paper, the GP covariance kernel is the squared exponential kernel $k(x,x')=\theta_0\exp\left(-\frac{\theta_1}{2}\|x-x'\|^2\right)$. The hyperparameters $\theta_0$ and $\theta_1$ can be optimized by performing maximization of $\mathcal{Q}$ over $\bm{\theta}=\{\theta_0,\theta_1\}$ using numerical packages. Normally, we update $\bm{\theta}$ every 20 iterations. 

Additional hyperparameters are the number and location of inducing points which affect the complexity and estimation quality of $\mu(t)$ and $\phi(\tau)$. A large number of inducing points will lead to high complexity while a small number cannot capture the dynamics. For fast inference, the inducing points are uniformly located on the domain. Another advantage of uniform location is that the kernel matrix has Toeplitz structure ~\citep{cunningham2008fast} which means the matrix inversion can be implemented more efficiently. The number of inducing points is gradually increased until no more significant improvement. The final pseudo code is provided in Alg.\ref{alg1}. 

\begin{algorithm}
\SetAlgoLined
\KwResult{$\mu(t)=\lambda^*_{\mu}\sigma(f(t))$, $\phi(\tau)=\lambda^*_{\phi}\sigma(g(\tau))$}
 Initialize hyperparameters and $\mathbf{X}$, $\lambda_\mu^*$, $\lambda_\phi^*$, $\bm{\omega}_{ii}$, $\bm{\omega}_{ij}$, $\mathbf{f}_{t_s}$, $\mathbf{g}_{\tau_s}$, $\Pi_\mu$, $\{\Pi_{\phi_i}\}_{i=1}^N$\;
 \For{}{
  Update the posterior of $\bm{\omega}_{ii}$ and $\bm{\omega}_{ij}$ by Eq.\eqref{eq11}\;
  Update intensities of $\Pi_\mu$ and $\{\Pi_{\phi}\}$ by Eq.\eqref{eq13}, \eqref{eq14}\;
  Update the posterior of $\mathbf{X}$ by Eq.\eqref{eq15}\;
  Update $\lambda^*_{\mu},\mathbf{f}_{t_s},\lambda^*_{\phi}$ and $\mathbf{g}_{\tau_s}$ by Eq.\eqref{eq20}\;
  Update hyperparameters. 
 }
 \caption{EM algorithm for SGPHP}
 \label{alg1}
\end{algorithm}

\section{Mean-field Variational Inference}
In the following, we extend the EM algorithm to a mean-field variational inference \citep{bishop2007pattern} algorithm which solves the inference problem slightly slower than EM, but can provide uncertainty with a distribution estimation rather than point estimation. 

In variational inference, the posterior distribution over latent variables is approximated by a variational distribution. The optimal variational distribution is chosen by minimising the Kullback-Leibler (KL) divergence or equivalently maximizing the evidence lower bound (ELBO). A common approach is the mean-field method where the variational distribution is assumed to factorize over some partition of latent variables. 

For the problem at hand, incorporating priors of $\lambda^*_{\mu}$, $f$, $\lambda^*_{\phi}$ and $g$ into Eq. \eqref{eq7} and \eqref{eq8}, we obtain the joint distribution over all variables. Without loss of generality, an improper prior $p(\lambda_{\cdot}^*)=1/\lambda_{\cdot}^*$ and a symmetric GP prior $\mathcal{GP}(\cdot|0,K_{\cdot})$ \citep{bishop2007pattern} are utilized here. We assume the variational distribution $q$ can factorize as
\begin{equation*}
\begin{aligned}
&q(\Pi_{\mu},\bm{\omega}_{ii},\{\Pi_{\phi_i}\}_{i=1}^N,\bm{\omega}_{ij},\mathbf{X},\lambda^*_{\mu},f,\lambda^*_{\phi},g)=\\
&q_1(\Pi_{\mu},\bm{\omega}_{ii},\{\Pi_{\phi_i}\}_{i=1}^N,\bm{\omega}_{ij},\mathbf{X})q_2(\lambda^*_{\mu},f,\lambda^*_{\phi},g).
\end{aligned}
\end{equation*}

The derivation of variational mean-field approach is shown in the Appendix~\ref{app4}. It is similar with that of EM algorithm. Here, we only show the final result. The optimal distribution for each factor maximizing the ELBO is given by 

\subsubsection{Optimal Density of P\'{o}lya-Gamma Variables}

\begin{equation}
\label{eq21}
\begin{aligned}
q_1(\bm{\omega}_{ii})&=\prod_{i=1}^Np_{\text{PG}}(\omega_{ii}|1,\tilde{f}(t_i))\\
q_1(\bm{\omega}_{ij})&=\prod_{i=2}^N\prod_{j=1}^{i-1}p_{\text{PG}}(\omega_{ij}|1,\tilde{g}(\tau_{ij})),
\end{aligned}
\end{equation}
where $\tilde{f}(t_i)=\sqrt{\mathbb{E}(f^2(t_i))}$ and $\tilde{g}(\tau_{ij})=\sqrt{\mathbb{E}(g^2(\tau_{ij}))}$ which can be computed utilizing $\mathbb{E}(C^2)=\mathbb{E}^2(C)+\text{Var}(C)$. 

\subsubsection{Optimal Marked Poisson Processes}

\begin{equation}
\label{eq22}
\begin{aligned}
\Lambda_\mu^1(t,\omega_\mu)&={\tilde{\lambda}_\mu^*}\sigma(-\tilde{f}(t))p_{\text{PG}}(\omega_\mu|1,\tilde{f}(t))e^{(\tilde{f}(t)-\bar{f}(t))/2}\\
\Lambda_\phi^1(\tau,\omega_\phi)&={\tilde{\lambda}_\phi^*}\sigma(-\tilde{g}(\tau))p_{\text{PG}}(\omega_\phi|1,\tilde{g}(\tau))e^{(\tilde{g}(\tau)-\bar{g}(\tau))/2},
\end{aligned}
\end{equation}
where $\tilde{\lambda}_\mu^*=e^{\mathbb{E}(\log \lambda_\mu^*)}$, $\bar{f}(t)=\mathbb{E}(f(t))$, $\tilde{\lambda}_\phi^*=e^{\mathbb{E}(\log \lambda_\phi^*)}$ and $\bar{g}(\tau)=\mathbb{E}(g(\tau))$. 

\subsubsection{Optimal Density of Intensity Upper-bounds}

\begin{equation}
\label{eq23}
\begin{aligned}
q_2(\lambda_\mu^*)&=\text{Gamma}(\lambda_\mu^*|\alpha_\mu,\beta_\mu)\\
q_2(\lambda_\phi^*)&=\text{Gamma}(\lambda_\phi^*|\alpha_\phi,\beta_\phi),
\end{aligned}
\end{equation}
where $\alpha_\mu=\sum_{i=1}^N \mathbb{E}(x_{ii})+\iint\Lambda_\mu^1(t,\omega_\mu)dtd\omega_\mu$, $\beta_\mu=T$, $\alpha_\phi=\sum_{i=1}^N\sum_{j=1}^{i-1}\mathbb{E}(x_{ij})+N\iint\Lambda_\phi^1(\tau,\omega_\phi)d\tau d\omega_\phi$, $\beta_\phi=NT_\phi$ and all intractable integrals can be solved by Gaussian quadrature. This provides the required expectation for Eq.\eqref{eq22} by $\mathbb{E}(\lambda^*)=\alpha/\beta$ and $\mathbb{E}(\log \lambda^*)=\psi(\alpha)-\log\beta$ where $\psi(\cdot)$ is the digamma function. 

\subsubsection{Optimal Sparse Gaussian Process}

\begin{equation}
\label{eq24}
\begin{aligned}
q_2(\mathbf{f}_{t_s})&=\mathcal{N}(\mathbf{f}_{t_s}|\tilde{\mathbf{m}}_{t_s},\tilde{\mathbf{\Sigma}}_{t_s})\\
q_2(\mathbf{g}_{\tau_s})&=\mathcal{N}(\mathbf{g}_{\tau_s}|\tilde{\mathbf{m}}_{\tau_s},\tilde{\mathbf{\Sigma}}_{\tau_s}),
\end{aligned}
\end{equation}
where $\tilde{\mathbf{\Sigma}}_{t_s}=\left[\mathbf{K}_{t_s t_s}^{-1}\int \tilde{A}_\mu(t)\mathbf{k}_{t_s t}\mathbf{k}_{t_s t}^Tdt\mathbf{K}_{t_s t_s}^{-1}+\mathbf{K}_{t_s t_s}^{-1}\right]^{-1}$, $\tilde{\mathbf{m}}_{t_s}=\tilde{\mathbf{\Sigma}}_{t_s}\mathbf{K}_{t_s t_s}^{-1}\int\tilde{B}_\mu(t)\mathbf{k}_{t_s t}dt$ with $\tilde{A}_{\mu}(t)=\sum_{i=1}^N\mathbb{E}[\omega_{ii}]\mathbb{E}[x_{ii}]\delta(t-t_i)+\int_0^\infty \omega_{\mu}\Lambda_{\mu}^1(t,\omega_{\mu})d\omega_{\mu}$ and $\tilde{B}_{\mu}(t)=\frac{1}{2}\sum_{i=1}^N\mathbb{E}[x_{ii}]\delta(t-t_i)-\frac{1}{2}\int_0^\infty\Lambda_{\mu}^1(t,\omega_{\mu})d\omega_{\mu}$; $\tilde{\mathbf{\Sigma}}_{\tau_s}=\left[\mathbf{K}_{\tau_s \tau_s}^{-1}\int \tilde{A}_\phi(\tau)\mathbf{k}_{\tau_s \tau}\mathbf{k}_{\tau_s \tau}^Td\tau\mathbf{K}_{\tau_s \tau_s}^{-1}+\mathbf{K}_{\tau_s \tau_s}^{-1}\right]^{-1}$, $\tilde{\mathbf{m}}_{\tau_s}=\tilde{\mathbf{\Sigma}}_{\tau_s}\mathbf{K}_{\tau_s \tau_s}^{-1}\int \tilde{B}_\phi(\tau)\mathbf{k}_{\tau_s \tau}d\tau$ with $\tilde{A}_{\phi}(\tau)=\sum_{i,j}\mathbb{E}[\omega_{ij}]\mathbb{E}[x_{ij}]\delta(\tau-\tau_{ij})+N\int_0^\infty \omega_{\phi}\Lambda_{\phi}^1(\tau,\omega_{\phi})d\omega_{\phi}$ and
$\tilde{B}_{\phi}(\tau)=\frac{1}{2}\sum_{i,j}\mathbb{E}[x_{ij}]\delta(\tau-\tau_{ij})-\frac{N}{2}\int_0^\infty\Lambda_{\phi}^1(\tau,\omega_{\phi})d\omega_{\phi}$. All intractable integrals are solved by Gaussian quadrature. Note also the similarity to EM algorithm in Eq.\eqref{eq20}. 

\subsubsection{Optimal Density of Branching Structure}

\begin{equation}
\label{eq25}
\begin{aligned}
q_1(x_{ii}=1)&=\frac{\tilde{\mu}(t_i)}{\tilde{\mu}(t_i)+\sum_{j=1}^{i-1}\tilde{\phi}(\tau_{ij})}\\
q_1(x_{ij}=1)&=\frac{\tilde{\phi}(\tau_{ij})}{\tilde{\mu}(t_i)+\sum_{j=1}^{i-1}\tilde{\phi}(\tau_{ij})}, 
\end{aligned}
\end{equation}
with $\tilde{\mu}(t_i)=\tilde{\lambda}_\mu^*e^{\mathbb{E}(\log\sigma(f(t_i)))}$, $\tilde{\phi}(\tau_{ij})=\tilde{\lambda}_\phi^*e^{\mathbb{E}(\log\sigma(g(\tau_{ij})))}$. The $\mathbb{E}(\log\sigma(\cdot))$ term can be solved by Gaussian quadrature. 

\subsubsection{Hyperparameters}
Similarly, the hyperparameters $\theta_0$ and $\theta_1$ can be optimized by performing maximization of ELBO over $\{\theta_0,\theta_1\}$ using numerical packages. The optimization of number and location of inducing points is same as EM algorithm. The final pseudo code is provided in Alg.\ref{alg2}. 
\begin{algorithm}
\SetAlgoLined
\KwResult{$\mu(t)=\lambda^*_{\mu}\sigma(f(t))$, $\phi(\tau)=\lambda^*_{\phi}\sigma(g(\tau))$}
 Initialize hyperparameters and variational distributions of $\mathbf{X}$, $\lambda_\mu^*$, $\lambda_\phi^*$, $\bm{\omega}_{ii}$, $\bm{\omega}_{ij}$, $\mathbf{f}_{t_s}$, $\mathbf{g}_{\tau_s}$, $\Pi_\mu$, $\{\Pi_{\phi_i}\}_{i=1}^N$\;
 \For{}{
  Update $q_1$ of $\bm{\omega}_{ii}$ and $\bm{\omega}_{ij}$ by Eq.\eqref{eq21}\;
  Update $\Lambda^1$ of $\Pi_\mu$ and $\{\Pi_{\phi}\}$ by Eq.\eqref{eq22}\;
  Update $q_2$ of $\lambda_\mu^*$ and $\lambda_\phi^*$ by Eq.\eqref{eq23}\;
  Update $q_2$ of $\mathbf{f}_{t_s}$ and $\mathbf{g}_{\tau_s}$ by Eq.\eqref{eq24}\;
  Update $q_1$ of $\mathbf{X}$ by Eq.\eqref{eq25}\;
  Update hyperparameters. 
 }
 \caption{Mean-field algorithm for SGPHP}
 \label{alg2}
\end{algorithm}

\begin{figure*}
\centering
\begin{subfigure}[t]{0.35\linewidth}
    \centering
    \includegraphics[width=\linewidth]{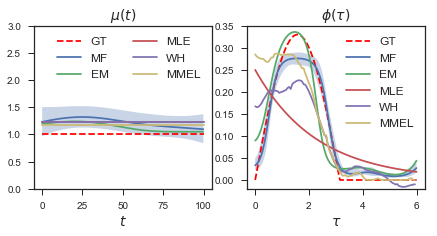}
    \caption{\label{figa}}
  \end{subfigure}
\begin{subfigure}[t]{0.35\linewidth}
    \centering
    \includegraphics[width=\linewidth]{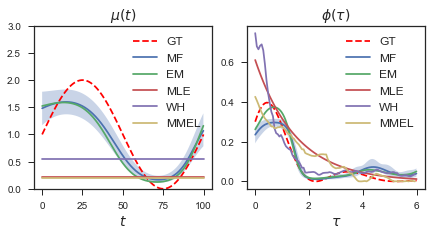}
    \caption{\label{figb}}
  \end{subfigure}
\begin{subfigure}[t]{0.19\linewidth}
    \centering
    \includegraphics[width=\linewidth]{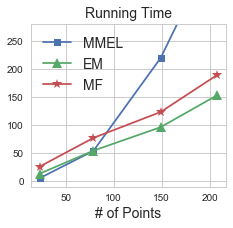}
    \caption{\label{figc}}
  \end{subfigure}
\caption{The fictitious data experimental results. (a): The estimated $\hat{\mu}(t)$ and $\hat{\phi}(\tau)$ in case 1 (with shading being one standard deviation for MF); (b): for case 2. We can see both EM and MF algorithms capture the structure of underlying rates better than other alternatives. (c): The running time (seconds) of different iterative nonparametric algorithms on varying $\#$ of observations. We can see EM and MF algorithms scale linearly with observation, which is more efficient than MMEL. (GT=Ground Truth)}
\label{fig1}
\end{figure*}

\section{Experiments}

We evaluate the performance of our proposed EM and mean-field (MF) algorithms on both fictitious and real-world data. Specifically, we compare our proposed algorithms to the following alternatives. \textbf{1. Maximum Likelihood Estimation (MLE)}: the vanilla Hawkes process with constant $\mu$ and exponential decay triggering kernel; \textbf{2. Wiener-Hopf (WH)}: a nonparametric algorithm for Hawkes process where $\mu$ is constant and $\phi(\tau)$ is a nonparametric function \citep{bacry2016first}; \textbf{3. Majorization Minimization Euler-Lagrange (MMEL)}: another nonparametric algorithm for Hawkes process with constant $\mu$ and smooth $\phi(\tau)$, which similarly utilizes the branching structure and estimates $\phi(\tau)$ by an Euler-Lagrange equation \citep{zhou2013learning}. We also tried to compare to the long short-term memory (LSTM) based neural Hawkes process \citep{mei2017neural} but found it hard to converge at least on our data. On the contrary, our proposed algorithms are easier to converge due to the fact that there are fewer parameters to tune, which constitutes another advantage. 

We use the following metrics to evaluate the performance
of various methods: \textbf{\textit{TestLL}}: the log-likelihood of hold-out data using the trained model. This is a metric describing the model prediction ability. \textbf{\textit{EstErr}}: the mean squared error between the estimated $\hat{\mu}(t)$, $\hat{\phi}(\tau)$ and the ground truth. It is only used for \textit{fictitious data}. \textbf{\textit{RunTime}}: the running time of various methods w.r.t. the number of training data. \textbf{\textit{Q-Q plot}}: the plot visualizes the goodness-of-fit for different models using time rescaling theorem \citep{papangelou1972integrability}. 

\begin{table}[ht]
\caption{\textit{EstErr} and \textit{TestLL} for fictitious and real datasets.}
\label{tab1}
\begin{center}
\scalebox{0.75}{
\begin{tabular}{c|c|c|c|c|c|c}
\hline
\multicolumn{2}{c|}{}   & MLE & WH & MMEL & EM & MF\\ \hline
\multirow{3}{*}{Case 1} & $\textit{EstErr}(\hat{\mu},\mu)$ & 0.236 & 0.228 & 0.173 & \textbf{0.137} & 0.223 \\ \cline{2-7} 
                        & $\textit{EstErr}(\hat{\phi},\phi)$ & 0.0289 & 0.0039 & 0.0053 & 0.0016 & \textbf{0.0005}\\ \cline{2-7}
                        & \textit{TestLL} & 31.21 & 33.72 & 32.78 & 33.87 & \textbf{33.98} \\ \hline
\multirow{3}{*}{Case 2} & $\textit{EstErr}(\hat{\mu},\mu)$ & 1.141 & 0.706 & 1.135 & 0.134 & \textbf{0.099}\\ \cline{2-7} 
                        & $\textit{EstErr}(\hat{\phi},\phi)$ & 0.0076 & 0.0086 & 0.0082 & \textbf{0.0011} & 0.0020 \\ \cline{2-7}
                        & \textit{TestLL} & 27.48 & 22.97 & 26.35 & \textbf{33.18} & 32.58\\ \hline
{Collision} & \textit{TestLL} & 420.41 & 439.67 & 470.56 & 470.34 & \textbf{494.46}\\ \hline
{Crime} & \textit{TestLL} & 371.59 & 400.36 & 381.42 & \textbf{520.22} & 375.29\\ \hline
\end{tabular}}
\end{center}
\end{table}

\subsection{Fictitious Data Experiments}
In fictitious data experiments, we use the thinning algorithm \citep{ogata1998space} to generate 100 sets of training data and 10 sets of hold-out data with $T_\phi=6$ and $T=100$ in 2 cases: \textbf{1.} $\mu(t)$ is a constant and \textbf{2.} $\mu(t)$ is changing over time. 

                    
\textbf{1.} $\mu(t)=1$ and $\phi(\tau)=\left\{
                        \begin{aligned}
                        &0.33\sin{\tau} \ \ (0<\tau\leq\pi)\\
                        &0 \ \ (\pi<\tau<T_\phi)
                        \end{aligned}
                    \right.$;
                    
\textbf{2.} $\mu(t)=\sin\left(\frac{2\pi}{T}\cdot t\right)+1 \ \ (0<t<T)$ and $\phi(\tau)=0.3\left(\sin(\frac{2\pi}{3}\cdot\tau)+1\right)\cdot\exp(-0.7\tau) \ \ (0<\tau<T_\phi)$. 

The inducing points and hyperparameters are optimized for inference. The estimated $\hat{\mu}(t)$ and $\hat{\phi}(\tau)$ are shown in Fig.\ref{figa} and \ref{figb}. From accuracy perspective, the result in Tab.\ref{tab1} confirms that \textbf{EM and MF algorithms outperform other alternatives in all cases w.r.t. \textit{TestLL} and \textit{EstErr}} (mean for MF). For MLE, the estimated results are far away from ground truth due to parametric constraint. For WH and MMEL, the constant limitation on $\mu(t)$ still exists even though $\phi(\tau)$ has been relieved. On the contrary, our EM and MF algorithms provide nonparametric $\hat{\mu}(t)$ and $\hat{\phi}(\tau)$ concurrently. From efficiency perspective, EM and MF algorithms are compared to the other iterative nonparametric algorithm MMEL with same iterations. We can see \textbf{EM and MF's \textit{RunTime} scales linearly with observation} in Fig.\ref{figc}, which proves their superior efficiency.

\subsection{Real Data Experiments}
In the real data section, we apply our EM and MF algorithms to two different datasets, 

\textbf{\textit{Motor Vehicle Collisions in New York City}}
: a vehicle collision dataset from New York City Police Department; 

\textbf{\textit{Crime in Vancouver}}: a dataset of crimes in Vancouver from the Vancouver Open Data Catalogue.


More experimental details (e.g. dataset introduction, data pre-processing, hyperparameter selection) are provided in Appendix~\ref{app5}. We compare the performance of our proposed algorithms to other alternatives. For each inference algorithm, we evaluate its predictive performance using \textit{TestLL}. The \textit{TestLL} of EM, MF and other alternatives are shown in Tab.\ref{tab1}. We can observe our EM and MF algorithms' consistent superiority over other alternatives whose baseline intensity or triggering kernel is too restricted to capture the dynamics. To further measure the performance, we generate the \textit{Q-Q plot}. We transform a sequence of timestamps in hold-out data by the fitted model to a set of independent uniform random variables on the interval $(0,1)$. The result is shown in Appendix~\ref{app5}. All experimental results prove \textbf{our proposed algorithms can not only describe $\mu(t)$ and $\phi(\tau)$ in a completely flexible manner which leads to better goodness-of-fit but also with superior efficiency}. 


\section{Conclusions}

In this paper, the scalable EM and mean-filed variational inference algorithms are proposed for sigmoid Gaussian Hawkes process. By augmenting the branching structure, P\'{o}lya-Gamma random variables and latent marked Poisson processes, the inference can be performed in a conjugate way efficiently. Furthermore, by introducing sparse GP approximation, our proposed algorithms scale linearly with the number of observations. The fictitious and real data experimental results confirm that the performance of accuracy and efficiency of our proposed algorithms is superior to other state-of-the-art alternatives. Future work can be done on the extension to multivariate or multidimensional Hawkes processes.

\bibliographystyle{aaai}
\bibliography{mybibfile}

\newpage
\appendix
\setcounter{secnumdepth}{0}
\section{Appendices}
\addcontentsline{toc}{section}{Appendices}
\renewcommand{\thesubsection}{\Roman{subsection}}
\setcounter{secnumdepth}{2}
\setcounter{equation}{0}
\subsection{Proof of Transformation of Sigmoid Function}
\label{app1}
\citet{polson2013bayesian} found that the inverse hyperbolic cosine can be expressed as an infinite mixture of Gaussian densities
\begin{equation}
\label{apeq1}
\cosh^{-b}{(z/2)}=\int_0^\infty e^{-z^2\omega/2}p_{\text{PG}}(\omega|b,0)d\omega,
\end{equation}
where $p_{\text{PG}}(\omega|b,0)$ is the P\'{o}lya-Gamma distribution with $\omega\in \mathbb{R}^+$. As a result, the sigmoid function can be defined as a Gaussian representation
\begin{equation}
\label{apeq2}
\sigma(z)=\frac{e^{z/2}}{2\cosh{(z/2)}}=\int_0^\infty e^{h(\omega,z)}p_{\text{PG}}(\omega|1,0)d\omega,
\end{equation}
where $h(\omega,z)=z/2-z^2\omega/2-\log2$. This proves Eq.\eqref{eq4} in the main paper. 

\subsection{Campbell's Theorem}
\label{app2}
Let $\Pi_{\hat{\mathcal{Z}}}=\{(\mathbf{z}_n,\bm{\omega}_n)\}_{n=1}^N$ be a marked Poisson process on the product space $\hat{\mathcal{Z}}=\mathcal{Z}\times \Omega$ with intensity $\Lambda(\mathbf{z},\bm{\omega})=\Lambda(\mathbf{z})p(\bm{\omega}|\mathbf{z})$. $\Lambda(\mathbf{z})$ is the intensity for the unmarked Poisson process $\{\mathbf{z}_n\}_{n=1}^N$ with $\bm{\omega}_n\sim p(\bm{\omega}_n|\mathbf{z}_n)$ being an independent mark drawn at each $\mathbf{z}_n$. Furthermore, we define a function $h(\mathbf{z},\bm{\omega}):\mathcal{Z}\times \Omega\rightarrow \mathbb{R}$ and the sum $H(\Pi_{\hat{\mathcal{Z}}})=\sum_{(\mathbf{z},\bm{\omega})\in\Pi_{\hat{\mathcal{Z}}}}h(\mathbf{z},\bm{\omega})$. If $\Lambda(\mathbf{z},\bm{\omega})<\infty$, then
\begin{equation*}
\begin{aligned}
&\mathbb{E}_{\Pi_{\hat{\mathcal{Z}}}}\left[\exp{\left(\xi H(\Pi_{\hat{\mathcal{Z}}})\right)}\right]\\
&=\exp{\left[\int_{\hat{\mathcal{Z}}}\left(e^{\xi h(\mathbf{z},\bm{\omega})}-1\right)\Lambda(\mathbf{z},\bm{\omega})d\bm{\omega}d\mathbf{z}\right]},
\end{aligned}
\end{equation*}
for any $\xi\in \mathbb{C}$. The above equation defines the characteristic functional of a marked Poisson process. This proves Eq.\eqref{eq6} in the main paper. The mean and variance are
\begin{equation*}
\begin{aligned}
\mathbb{E}_{\Pi_{\hat{\mathcal{Z}}}}\left[H(\Pi_{\hat{\mathcal{Z}}})\right]&=\int_{\hat{\mathcal{Z}}}h(\mathbf{z},\bm{\omega})\Lambda(\mathbf{z},\bm{\omega})d\bm{\omega}d\mathbf{z}\\
\text{Var}_{\Pi_{\hat{\mathcal{Z}}}}\left[H(\Pi_{\hat{\mathcal{Z}}})\right]&=\int_{\hat{\mathcal{Z}}}[h(\mathbf{z},\bm{\omega})]^2\Lambda(\mathbf{z},\bm{\omega})d\bm{\omega}d\mathbf{z}
\end{aligned}
\end{equation*}

\subsection{Proof of Augmented Likelihood}
\label{app3}
Substituting Eq.\eqref{eq4} and \eqref{eq6} into Eq.\eqref{eq3} in the main paper, we obtain the augmented joint likelihood of baseline intensity part
\begin{equation*}
\begin{aligned}
&p(D,\mathbf{X}_{ii}|\lambda^*_{\mu},f)\\
&=\prod_{i=1}^N\left(\lambda^*_{\mu}\sigma(f(t_i))\right)^{x_{ii}}\exp{\left(-\int_{T}\lambda^*_{\mu}\sigma(f(t))dt\right)}\\
&=\prod_{i=1}^N\left(\int_0^\infty\lambda^*_{\mu} e^{h(\omega_{ii},f(t_i))}p_{\text{PG}}(\omega_{ii}|1,0)d\omega_{ii}\right)^{x_{ii}}\cdot\\
&\mathbb{E}_{p_{\lambda_\mu}}\left[\prod_{(\omega_{\mu},t)\in\Pi_{\mu}}e^{h(\omega_{\mu},-f(t))}\right]\\
&=\iint\prod_{i=1}^N\left(\lambda_{\mu}(t_i,\omega_{ii})e^{h(\omega_{ii},f(t_i))}\right)^{x_{ii}}\\
&\cdot p_{\lambda_\mu}(\Pi_\mu|\lambda^*_{\mu})\prod_{(\omega_{\mu},t)\in \Pi_\mu}e^{h(\omega_\mu,-f(t))}d\bm{\omega}_{ii}d\Pi_\mu.
\end{aligned}
\end{equation*}
with $\bm{\omega}_{ii}$ denoting a vector of $\omega_{ii}$ and $\lambda_{\mu}(t_i,\omega_{ii})=\lambda^*_{\mu}p_{\text{PG}}(\omega_{ii}|1,0)$. 
Therefore, the augmented joint likelihood is 
\begin{equation*}
\begin{aligned}
&p(D,\Pi_{\mu},\bm{\omega}_{ii},\mathbf{X}_{ii}|\lambda^*_{\mu},f)=\prod_{i=1}^N\left(\lambda_{\mu}(t_i,\omega_{ii})e^{h(\omega_{ii},f(t_i))}\right)^{x_{ii}}\\
&\cdot p_{\lambda_\mu}(\Pi_\mu|\lambda^*_{\mu})\prod_{(\omega_{\mu},t)\in \Pi_\mu}e^{h(\omega_\mu,-f(t))}
\end{aligned}
\end{equation*}
This proves Eq.\eqref{eq7} in the main paper. The proof of Eq.\eqref{eq8} in the main paper is the same and omitted here.

\subsection{Derivation of Mean-field Approach}
\label{app4}
A standard derivation in the variational mean–field approach shows that the optimal distribution for each factor maximizing the ELBO is given by 
\begin{equation}
\label{apeq3}
\begin{aligned}
\log &q_1(\Pi_{\mu},\bm{\omega}_{ii},\{\Pi_{\phi_i}\}_{i=1}^N,\bm{\omega}_{ij},\mathbf{X})=\\
&\mathbb{E}_{q_2}[\log p(\Pi_{\mu},\bm{\omega}_{ii},\{\Pi_{\phi_i}\}_{i=1}^N,\bm{\omega}_{ij},\mathbf{X},\lambda^*_{\mu},f,\lambda^*_{\phi},g)]+C_1\\
\log &q_2(\lambda^*_{\mu},f,\lambda^*_{\phi},g)=\\
&\mathbb{E}_{q_1}[\log p(\Pi_{\mu},\bm{\omega}_{ii},\{\Pi_{\phi_i}\}_{i=1}^N,\bm{\omega}_{ij},\mathbf{X},\lambda^*_{\mu},f,\lambda^*_{\phi},g)]+C_2
\end{aligned}
\end{equation}

For our problem, incorporating priors of $\lambda^*_{\mu}$, $f$, $\lambda^*_{\phi}$ and $g$ into Eq. \eqref{eq7} and \eqref{eq8} in the main paper, we obtain the joint distribution over all variables. Without loss of generality, an improper prior $p(\lambda_{\cdot}^*)=1/\lambda_{\cdot}^*$ and a symmetric GP prior $\mathcal{GP}(\cdot|0,K_{\cdot})$ \citep{bishop2007pattern} are utilized here. The final augmented joint distribution of baseline intensity part is
\begin{equation}
\label{apeq4}
\begin{aligned}
p(D,&\Pi_{\mu},\bm{\omega}_{ii},\mathbf{X}_{ii},\lambda^*_{\mu},f)=\prod_{i=1}^N\left(\lambda_{\mu}(t_i,\omega_{ii})e^{h(\omega_{ii},f(t_i))}\right)^{x_{ii}}\\
&\cdot p_{\lambda_\mu}(\Pi_\mu|\lambda^*_{\mu})\prod_{(\omega_{\mu},t)\in \Pi_\mu}e^{h(\omega_\mu,-f(t))}\cdot{\lambda_\mu^*}^{-1}\mathcal{GP}(f);
\end{aligned}
\end{equation}
and that of triggering kernel part is
\begin{equation}
\label{apeq5}
\begin{aligned}
&p(D,\{\Pi_{\phi_i}\}_{i=1}^N,\bm{\omega}_{ij},\mathbf{X}_{ij},\lambda^*_{\phi},g)=\\
&\prod_{i=2}^N\prod_{j=1}^{i-1}\left(\lambda_{\phi}(\tau_{ij},\omega_{ij})e^{h(\omega_{ij},g(\tau_{ij}))}\right)^{x_{ij}}\cdot\\
&\prod_{i=1}^N\left[p_{\lambda_\phi}(\Pi_{\phi_i}|\lambda^*_{\phi})\prod_{(\omega_{\phi},\tau)\in \Pi_{\phi_i}}e^{h(\omega_\phi,-g(\tau))}\right]\cdot{\lambda_\phi^*}^{-1}\mathcal{GP}(g).
\end{aligned}
\end{equation}

Substituting Eq.\eqref{apeq4} and \eqref{apeq5} into Eq.\eqref{apeq3}, we can obtain the optimal distribution for each factor (Eq.\eqref{eq21}$\sim$\eqref{eq25} in the main paper). What is worth noting is that $\bm{\omega}_{ii}$ and $\bm{\omega}_{ij}$ are coupled with branching structure $\mathbf{X}$. We marginalize out $\bm{\omega}$ when solving $q_1(\mathbf{X})$ and vice versa. 

\subsection{Experimental Details}
\label{app5}
In this appendix, we elaborate on the details of data generation, processing, hyperparameter setup and some experimental results. 

\subsubsection{Fictitious Data Experimental Details}

The first case has a constant $\mu(t)$ and half sinusoidal triggering kernel $\phi(\tau)$. The bandwidth of WH is set to 0.3 and there are 10 inducing points on both $\mu(t)$ and $\phi(\tau)$ for EM and MF. The learned results in Fig.\ref{fig1} and Tab.\ref{tab1} prove EM and MF can learn the correct triggering kernel in non-monotonic decreasing cases. 

The second case is a general case with time-changing $\mu(t)$ and sinusoidal exponential decay triggering kernel. The bandwidth of WH is 0.1 and there are 10 inducing points on both $\mu(t)$ and $\phi(\tau)$ for EM and MF. Learned results in Fig.\ref{fig1} and Tab.\ref{tab1} show our proposed algorithms are still the champion. 

\subsubsection{Vehicle Collision Dataset} In daily transportation, car collisions happening in the past will have a triggering influence on the future because of the traffic congestion caused by the initial accident, so there is a self-exciting phenomenon in this kind of application. The vehicle collision dataset is from New York City Police Department. We filter out weekday records in nearly one month (Sep.18th-Oct.13th 2017). The number of collisions in each day is about 600. Records in Sep.18th-Oct.6th are used as training data and Oct.9th-13th are held out as test data. 

We compare the performance of EM, MF and other alternatives. The whole observation period $T$ is set to 1440 minutes (24 hours) and the support of triggering kernel $T_\phi$ is set to 60 minutes. For hyperparameters, the bandwidth of WH is set to 0.3 and there are 10 inducing points on $\mu(t)$ and $\phi(\tau)$ for EM and MF with hyperparameters $\{\theta_0, \theta_1\}$ optimized for inference. The final result is the average of learned $\mu(t)$ or $\phi(\tau)$ of each day. 

\subsubsection{Crime in Vancouver} In crime domain, the past crime will have a triggering effect on future ones, which has been reported in a lot of literature. The data of crimes in Vancouver comes from the Vancouver Open Data Catalogue. It was extracted on 2017-07-18 and it includes all valid felony, misdemeanour and violation crimes from 2003-01-01 to 2017-07-13. The columns are crime type, year, month, day, hour, minute, block, neighbourhood, latitude, longitude etc. We filter out the crime records from 2016-06-01 to 2016-08-31 as training data and 2016-09-01 to 2016-11-30 as test data, add a small time interval to separate all the simultaneous records and delete some records with empty values. 

The whole observation period $T$ is set to 92 days and the support of $\phi(\tau)$ is set to 6 days. For hyperparameters, the bandwidth of WH is set to 0.5 and there are still 10 inducing points on $\mu(t)$ and $\phi(\tau)$ for EM and MF with hyperparameters $\{\theta_0, \theta_1\}$ optimized for inference. 

\subsubsection{Q-Q plot}
\begin{figure}
\centering
\begin{subfigure}[t]{0.49\linewidth}
    \centering
    \includegraphics[width=\linewidth]{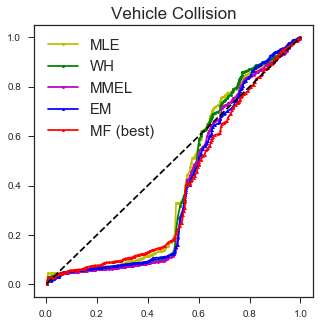}
    \caption{\label{apfiga}}
  \end{subfigure}
\begin{subfigure}[t]{0.49\linewidth}
    \centering
    \includegraphics[width=\linewidth]{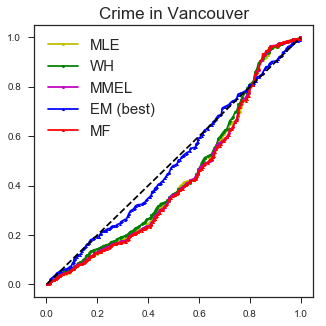}
    \caption{\label{apfigb}}
  \end{subfigure}
\caption{\textit{Q-Q plot} of EM, MF and other alternative inference algorithms for the real data.}
\label{apfig1}
\end{figure}

After the baseline intensity and triggering kernel estimated from the training data series, we can compute the rescaled timestamps of test data:
\begin{equation*}
\begin{aligned}
\tau_i=\Lambda(t_i)-\Lambda(t_{i-1}),
\end{aligned}
\end{equation*}
where $\Lambda(t_i)=\int_0^{t_i}\lambda(t|\mathcal{H}_{t_i})dt$, $\mathcal{H}_{t_i}$ is the history before $t_i$. According to the time rescaling theorem, $\{\tau_i\}$ are independent exponential random variables with mean $1$ if the model is correct. With further transformation:
\begin{equation*}
\begin{aligned}
z_i=1-\exp(-\tau_i),
\end{aligned}
\end{equation*}
$\{z_i\}$ are independent uniform random variables on the interval $(0,1)$. Any statistical assessment that measures agreement between the transformed data and a uniform distribution evaluates how well the fitted model agrees with the test data. So we can draw the \textit{Q-Q plot} of the transformed timestamps with respect to the uniform distribution. The plots are shown in Fig.~\ref{apfig1}. The perfect model follows a straight line of $y = x$. We can observe that the learned model from our algorithm (the better one from EM or MF) is generally closer to the straight line, which suggests its better goodness-of-fit. 

\end{document}